\documentclass[letterpaper, 10pt, journal, twoside]{IEEEtran}
\usepackage[utf8]{inputenc} 
\usepackage[english]{babel}

\usepackage[lined,ruled]{algorithm2e}
\usepackage{amsmath, amssymb, amsfonts}
\usepackage{bm}
\usepackage{booktabs}
\usepackage{cite}  
\usepackage{color}
\usepackage{graphicx}
\usepackage{siunitx}
\usepackage{subcaption}
\usepackage{url}
\usepackage{hyperref}

\graphicspath{{./figures/}}

\title{Learning Vision-based Flight in Drone \\Swarms by Imitation}
\author{
Fabian Schilling, Julien Lecoeur, Fabrizio Schiano, and Dario Floreano%
\thanks{Manuscript received: May 27, 2019; Accepted July 28, 2019.}
\thanks{This paper was recommended for publication by Editor Jonathan Roberts upon evaluation of the Associate Editor and Reviewers' comments.
This work was supported by the Swiss National Science Foundation (SNSF) with grant number 200021-155907 and the European project RoboCom++.}
\thanks{The authors are with the Laboratory of Intelligent Systems (LIS), École Polytechnique Fédérale de Lausanne (EPFL), 1015 Lausanne, Switzerland. {\tt\footnotesize \href{https://lis.epfl.ch}{https://lis.epfl.ch}}}
\thanks{Corresponding author: {\tt\footnotesize \href{mailto:fabian.schilling@epfl.ch}{fabian.schilling@epfl.ch}}}
\thanks{\copyright~2019 IEEE. Personal use is permitted, but republication/redistribution requires IEEE permission.}
}

\newcommand{\tit}[1]{\textit{#1}}

\newcommand{\tno}[1]{\textnormal{#1}}

\newcommand{\mbf}[1]{\mathbf{#1}}
\newcommand{\mcl}[1]{\mathcal{#1}}
\newcommand{\mbb}[1]{\mathbb{#1}}
\newcommand{\mtx}[1]{\text{#1}}
\newcommand{\txt}[1]{\text{#1}}

\newcommand{\Vector}[1]{\mbf{#1}}
\newcommand{\vel}{\Vector{v}}
\newcommand{\pos}{\Vector{p}}
\newcommand{\rel}{\Vector{r}}

\newcommand{\World}{\mcl{W}}
\newcommand{\Body}{\mcl{B}}

\newcommand{\SO}[1]{\txt{SO}(#1)}

\newcommand{\Rotation}{\mbf{R}}

\newcommand{\policy}{\pi}
\newcommand{\dataset}{\mcl{D}}
\newcommand{\obs}{\mbf{o}}
\newcommand{\state}{\mbf{s}}
\newcommand{\action}{\mbf{a}}
\newcommand{\expert}{\policy^*}
\newcommand{\learner}{\hat{\policy}}

\newcommand{\norm}[1]{\|#1\|}

\newcommand{\set}[1]{\mcl{#1}}

\newcommand{\neighbors}{\set{N}}

\newcommand{\Real}{\mbb{R}}

\newcommand{\target}[1]{#1}

\begin{document}

\maketitle


\begin{abstract}
Decentralized drone swarms deployed today either rely on sharing of positions among agents or detecting swarm members with the help of visual markers.
This work proposes an entirely visual approach to coordinate markerless drone swarms based on imitation learning.
Each agent is controlled by a small and efficient convolutional neural network that takes raw omnidirectional images as inputs and predicts 3D velocity commands that match those computed by a flocking algorithm.
We start training in simulation and propose a simple yet effective unsupervised domain adaptation approach to transfer the learned controller to the real world.
We further train the controller with data collected in our motion capture hall.
We show that the convolutional neural network trained on the visual inputs of the drone can learn not only robust inter-agent collision avoidance but also cohesion of the swarm in a sample-efficient manner.
The neural controller effectively learns to localize other agents in the visual input, which we show by visualizing the regions with the most influence on the motion of an agent.
We remove the dependence on sharing positions among swarm members by taking only local visual information into account for control.
Our work can therefore be seen as the first step towards a fully decentralized, vision-based swarm without the need for communication or visual markers.
\end{abstract}



\section*{Supplementary Material}\label{sec:supplementary-material}

Supplementary video: \href{https://youtu.be/I9vFvPphfpU}{https://youtu.be/I9vFvPphfpU}.


\section{Introduction}\label{sec:introduction}

\IEEEPARstart{C}{ollective} motion of animal groups such as flocks of birds is an awe-inspiring natural phenomenon that has profound implications for the field of aerial swarm robotics \cite{floreano_science_2015,zufferey_aerial_2011}.
Animal groups in nature operate in a completely self-organized manner since the interactions between them are purely local.
By taking inspiration from decentralization in biological systems, we can develop powerful robotic swarms that are 1) robust to failure, and 2) highly scalable since the number of agents can be increased or decreased dynamically depending on the workload of the task.

One of the most appealing characteristics of collective animal behavior for robotics is that decisions are made based on local information.
Thus, the behavior of animal groups does not require extensive knowledge of the swarm state or a central coordinator.
As of today, however, most multi-agent robotic systems rely on entirely centralized control \cite{kushleyev_towards_2013,preiss_crazyswarm_2017,weinstein_visual_2018} or wireless communication of positions \cite{vasarhelyi_outdoor_2014,viragh_flocking_2014,vasarhelyi_optimized_2018}, which are obtained either from a motion capture system or global navigation satellite system (GNSS).
The main drawback of these approaches is the introduction of a single point of failure, as well as the use of unreliable data links, respectively.
Relying on centralized control bears a significant risk since the agents lack the autonomy to make their own decisions in failure cases such as a communication outage.
The possibility of failure is even higher in dense urban environments, where GNSS measurements are often unreliable and imprecise.

\begin{figure}[t!]
    \centering
    \includegraphics[width=\columnwidth]{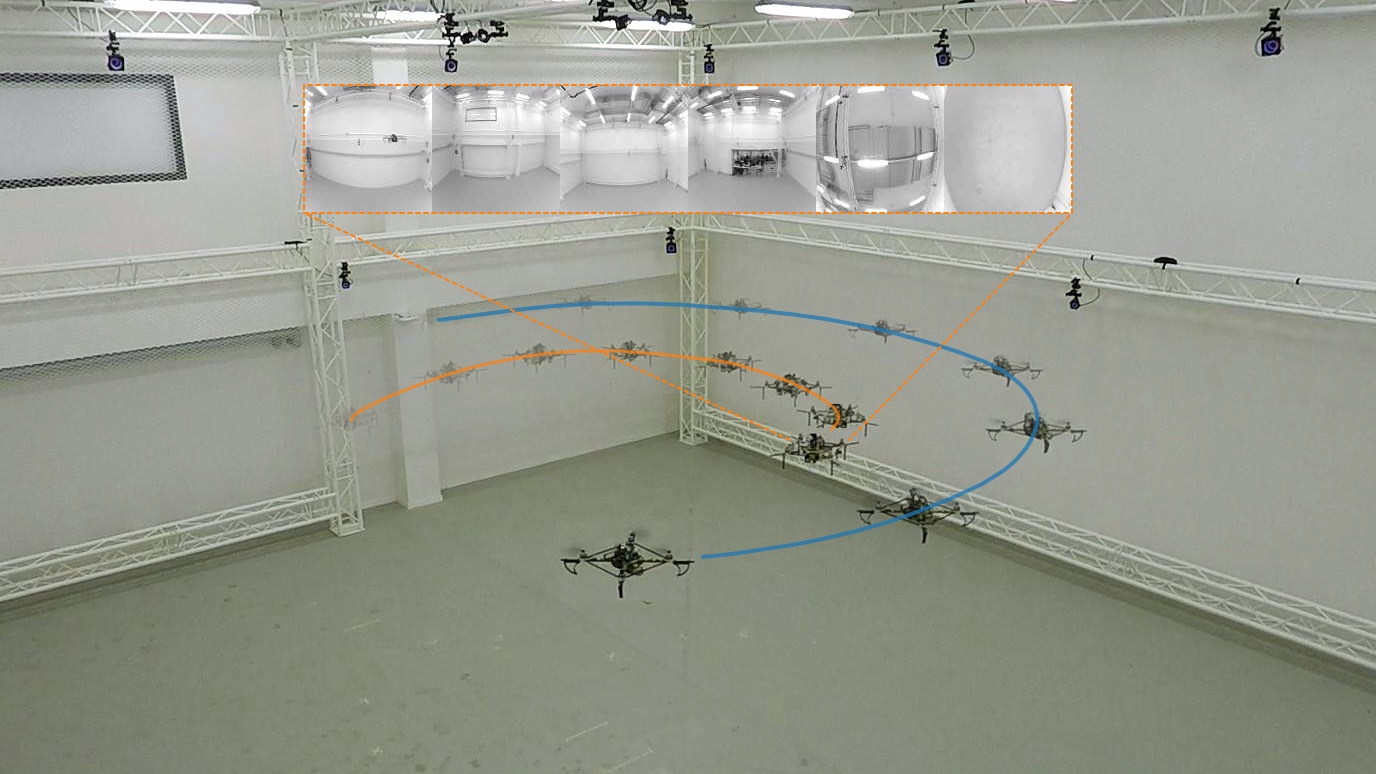}
    \caption{%
        Vision-based multi-agent experiment in our motion tracking hall.
        Our proposed visual controller operates fully decentralized and provides collision-free, coherent collective motion without the need to share positions among agents.
        The behavior of an agent only depends on its omnidirectional visual inputs (see orange rectangle).
        Collision avoidance and cohesion between agents are learned entirely from visual inputs (see supplementary video).
    }\label{fig:overview}
\end{figure}

Vision is arguably the most promising sensory modality to achieve a maximum level of autonomy for robotic systems, particularly considering the recent advances in computer vision and deep learning \cite{lecun_deep_2015}.
Apart from being light-weight and having relatively low power consumption, even cheap commodity cameras provide an unparalleled information density with respect to sensors of similar cost.
Their characteristics are specifically desirable for the deployment of an aerial multi-robot system.
The difficulty when using cameras for robot control is the processing of the visual information which this paper addresses directly.

In this work, we propose a reactive control strategy based entirely on local visual information.
We formulate the swarm interactions as a regression problem in which we predict control commands as a nonlinear function of the visual input of a single agent.
To the best of our knowledge, this is the first successful attempt to learn vision-based swarm behaviors such as collision-free navigation in an end-to-end manner directly from raw images.

Our contributions can be summarized as follows:
\begin{itemize}
    \item We propose a data-efficient imitation learning approach to solve the problem of vision-based coordination of a swarm of drones.
    Our control policy is trained incrementally by following the previous best policy and thus collecting relevant data from its failure cases.
    Our proposed system generates high-level control commands from raw images in the form of velocity setpoints, whereas a classical cascaded feedback control architecture handles low-level control.
    \item We present a remarkably simple and effective task-specific unsupervised domain adaptation approach to transfer the image data obtained from simulation to the real world.
    To this end, we collect a dataset of unlabeled images from our target environment to serve as backgrounds for images generated in simulation.
    \item We implement our algorithm on a physical quadrotor platform and show that all computations (policy evaluation, state estimation, and control) can be run entirely onboard in real-time.
    \item We provide an evaluation of our system in simulation and experimental validation in a motion tracking hall to show that our control policy generalizes to coordinated multi-agent flights in the real world.
\end{itemize}


\section{Related Work}\label{sec:related-work}


Decentralized swarms of drones such as quadrotors and fixed-wings are the focus of recent research in swarm robotics.
Early work presents ten fixed-wing drones deployed in an outdoor environment \cite{hauert_reynolds_2011}.
Their collective motion is based on Reynolds flocking \cite{reynolds_flocks_1987} with a migration term that allows the swarm to navigate towards the desired goal.
Thus far, the largest decentralized quadrotor swarm consisted of 30 autonomous agents flying in an outdoor environment \cite{vasarhelyi_optimized_2018}.
The underlying algorithm has many free parameters which are optimized using an evolutionary algorithm that relies on a fitness function which incorporates several swarm order parameters.
The commonality of the mentioned approaches and others, for example, \cite{vasarhelyi_outdoor_2014,dousse_human-comfortable_2017}, is the ability to share GNSS positions wirelessly among swarm members.
However, there are many situations in which wireless communication is unreliable, or GNSS positions are too imprecise.
We may not be able to tolerate position imprecisions in situations where the environment requires a small inter-agent distance, for example when traversing narrow passages in urban environments.
In these situations, tall buildings may deflect the signal and communication outages occur due to the wireless bands being over-utilized.


Recent advances in the field of machine learning facilitate vision-based control of flying robots.
In particular, the controllers are based on three types of learning methods: imitation learning, supervised learning, and reinforcement learning.
Imitation learning is used in \cite{ross_learning_2013} to control a drone in a forest environment based on human pilot demonstrations.
The authors motivate the importance of following suboptimal control policies in order to cover more of the state space.
A supervised learning approach \cite{loquercio_dronet_2018} features a convolutional network that is used to predict a steering angle and a collision probability for drone navigation in urban environments.
In contrast with the previous methods based only on supervised learning, an approach based on reinforcement learning \cite{sadeghi_cad2rl_2017} shows that a neural network trained entirely in a simulated environment can generalize to flights in the real world.
The work described above and other similar methods, for instance, \cite{giusti_machine_2016,smolyanskiy_toward_2017}, use a data-driven approach to control a flying robot in real-world environments.
The probability of collision is learned by minimizing the binary cross-entropy of labeled images collected while riding a bicycle through urban environments.
A shortcoming of these methods is that the learned controllers operate only in two-dimensional space which bears similar characteristics to navigation with ground robots.
Moreover, the approaches do not show the ability of the controllers to coordinate a multi-agent system.


The control of multiple agents based on visual inputs is achieved with relative localization techniques \cite{saska_system_2017} for a group of three quadrotors.
Each agent is equipped with a camera and a circular marker that enables the detection of other agents and the estimation of relative distance.
The system relies only on local information obtained from the onboard cameras in near real-time.
Thus far, decentralized vision-based drone control has been realized by mounting visual markers on the drones \cite{krajnik_practical_2014}.
Although this simplifies the relative localization problem significantly, the marker-based approach would not be desirable for real-world deployment of flying robots.
The used visual markers are relatively large and bulky which unnecessarily adds weight and drag to the platform; this is especially detrimental in real-world conditions.
Another recently proposed approach is the use of active ultraviolet markers to identify the relative range and bearing to other agents \cite{walter_uvdar_2019}.
However, the markers have to be placed in carefully chosen pre-defined locations, and the system is thus unable to detect markerless drones that do not precisely conform to these specifications.


\section{Method}\label{sec:method}

At the core of our method lies the prediction of a velocity command for each agent that matches the velocity command computed by a flocking algorithm.
We consider the velocity command from the flocking algorithm as the target for a supervised imitation learning problem.
The main idea is to eliminate the dependence on the knowledge of the positions of other agents by taking only local visual information into account for control.
Imitation learning presents a practical alternative to the approach in which separate modules are responsible for object detection, multi-object target tracking, and control, respectively.
The modular approach would require the manual labeling of prohibitively large amounts of images with precise bounding box annotations.
By using direct imitation, the control inputs can be calculated directly from the relative positions of other agents obtained either from simulation or a motion capture system.

\subsection{Flocking algorithm}\label{sec:flocking-algorithm}

We use an adaptation of Reynolds flocking \cite{reynolds_flocks_1987} to generate targets for our learning algorithm.
In particular, we only consider the \tit{collision avoidance} and \tit{flock centering} terms from the original formulation since they only depend on relative positions.
We omit the \tit{velocity matching} term since estimating the velocities of other agents is a challenging task given only a single snapshot in time.
One would have to rely on either estimating velocities from several consecutive images or estimating the orientation and heading with relatively high precision in order to infer velocities from a single image.

In our formulation of the flocking algorithm, we use the terms \tit{separation} and \tit{cohesion} to denote collision avoidance and flock centering, respectively \cite{saska_swarms_2014}.
We further add an optional \tit{migration} term that enables the agents to navigate towards a goal.
An important consideration when modeling the desired behavior of the swarm is the notion of neighbor selection.
It is reasonable to assume that each agent can only perceive its neighbors in a limited range.
We therefore only consider agents as neighbors if they are closer than the desired cutoff distance $r^\txt{max}$ which corresponds to only selecting agents in a sphere with a given radius.
We do not make any restrictions on the field of view of the agents since limiting perception, specifically in the lateral direction, has been shown to have adverse effects on the flocking performance \cite{soria_influence_2019}.
Therefore, we denote the set of neighbors of an agent $i$ as the set
$ \neighbors_i = \left\{ \txt{agents}~j : j \neq i \wedge \norm{\rel_{ij}} < r^\txt{max} \right\} $
where $\rel_{ij} \in \Real^3$ denotes the relative position of agent $j$ with respect to agent $i$ and $\norm{\cdot}$ the Euclidean norm.
We compute $\rel_{ij} = \pos_j - \pos_i$ where $\pos_i \in \Real^3$ denotes the absolute position of agent $i$.

The separation term steers an agent away from its neighbors in order to avoid collisions, whereas the cohesion term can be seen as the antagonistic inverse since its purpose is to steer an agent towards its neighbors to provide coherence to the group.
We can formalize the respective separation and cohesion velocity command for the $i$th agent as

\begin{align}\label{eq:separation-cohesion}
    \vel_i^\txt{sep} &= -\frac{k^\txt{sep}}{|\neighbors_i|} \sum_{j \in \neighbors_i} \frac{\rel_{ij}}{\norm{\rel_{ij}}^2} \\
    \vel_i^\txt{coh} &= \frac{k^\txt{coh}}{|\neighbors_i|} \sum_{j \in \neighbors_i} \rel_{ij}
\end{align}
where $k^\txt{sep}$ is the separation gain which modulates the strength of the separation between agents and the cohesion gain $k^\text{coh}$ modulates the tendency for the agents to be drawn towards the center of the neighboring agents.
For our implementation, the separation and cohesion terms are sufficient to generate a collision-free swarm in which agents remain together, given that the separation and cohesion gains are chosen carefully.
We denote the combination of the two terms as the Reynolds velocity command $\vel_i^\txt{rey} = \vel_i^\txt{sep} + \vel_i^\txt{coh}$ which is later predicted by the neural network.

Moreover, the addition of the migration term provides the possibility to give a uniform navigation goal to all agents.
The corresponding migration velocity command is given by

\begin{equation}
    \vel_i^\txt{mig} = k^\txt{mig} \frac{\rel_i^\txt{mig}}{\norm{\rel_i^\txt{mig}}}
\end{equation}
where $k^\text{mig}$ denotes the migration gain and $\rel_i^\txt{mig} \in \Real^3$ denotes the relative position of the migration point with respect to agent $i$.
We compute $\rel_i^\txt{mig} = \pos^\txt{mig} - \pos_{i}$ where $\pos^\txt{mig} \in \Real^3$ is the absolute position of the migration point.

The velocity command for an agent $i$ is computed as a sum of the Reynolds terms, which is a combination of separation and cohesion, as well as the migration term, as $\tilde{\vel}_i = \vel_i^\txt{rey} + \vel_i^\txt{mig}$.
In general, we assume a homogeneous swarm, which means that all agents are given the same gains for separation, cohesion, and migration.

A final parameter to adjust the behavior of the swarm is the cutoff of the maximum speed.
The final velocity command that steers an agent is given by
$ \vel_i = \tilde{\vel}_i / \norm{\tilde{\vel}_i} \min \left( \norm{\tilde{\vel}_i}, v^\txt{max}\right) $
where $v^{\txt{max}}$ denotes the desired maximum speed of an agent during flocking.

\subsection{Drone model}\label{sec:drone-model}

\begin{figure}[t]
\centering
\begin{subfigure}[b]{\columnwidth}
    \centering
    \includegraphics[width=0.96\textwidth]{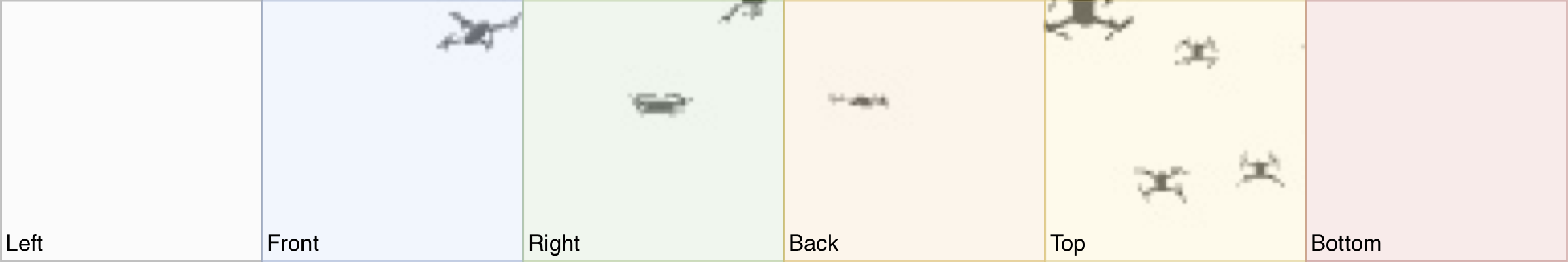}
    \caption{Visual input of an agent: concatenation of six camera images}\label{fig:field-of-view}
\end{subfigure}
\par\bigskip
\begin{subfigure}[b]{0.42\columnwidth}
    \centering
    \includegraphics[width=\textwidth]{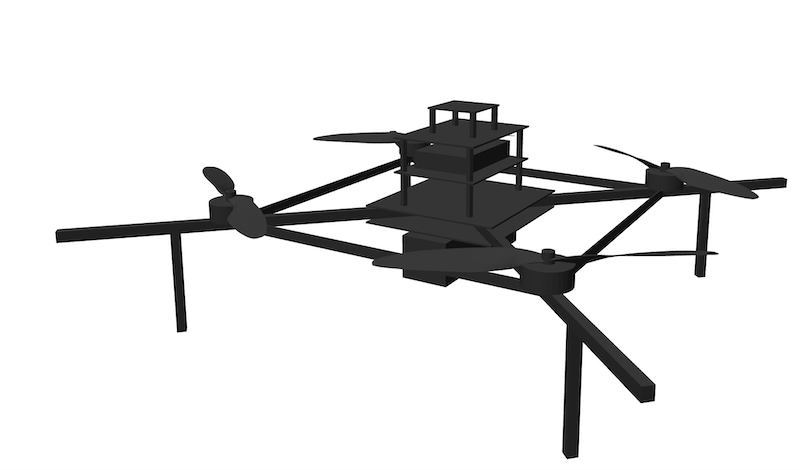}
    \caption{Simulated drone model}\label{fig:drone-model}
\end{subfigure}
\quad
\begin{subfigure}[b]{0.42\columnwidth}
    \centering
    \includegraphics[width=\textwidth]{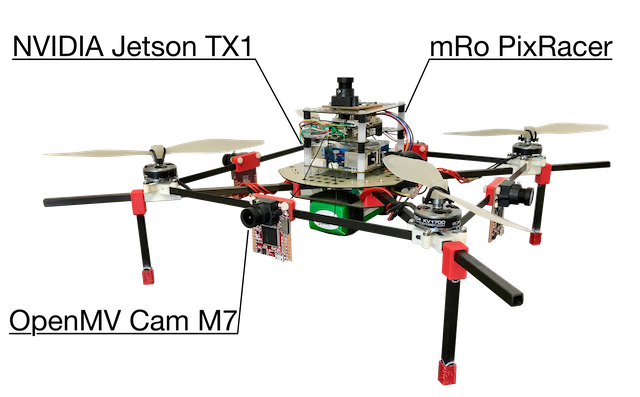}
    \caption{Physical drone hardware}\label{fig:drone-hardware}
\end{subfigure}
\caption{%
    Camera configuration and resulting visual input for a simulated agent.
    The cameras are positioned such that the visual field of an agent corresponds to an image cube map, i.e., each camera is pointing at a different face of a cube as seen from within the cube itself.
    (\ref{fig:field-of-view}) The concatenation of the six camera images into the full visual field (color-coded by camera).
    (\ref{fig:drone-model}) Simulated model of the drone built using mainly geometric primitives.
    (\ref{fig:drone-hardware}) Hardware implementation of the drone based on the design in \cite{dousse_human-comfortable_2017}.
    It uses six OpenMV Cam M7 with ultra-wide angle lenses for image acquisition, an NVIDIA Jetson TX1 for image processing, and a Pixracer autopilot for state estimation and control.
    }\label{fig:camera-config}
\end{figure}

We perform simulation in Gazebo with a group of nine quadrotor drones, each equipped with six simulated cameras to provide omnidirectional vision.
The cameras are positioned away from the center of gravity of the drone in order to have an unobstructed view of the surrounding environment, including the propellers (see Fig.~\ref{fig:field-of-view}).
Each camera has a $135 \times 90^\circ$ horizontal and vertical field of view and takes a grayscale image of $128 \times 128$ pixels with a refresh rate of $10$ Hz.
We concatenate the images from all six cameras along the horizontal axis to form a $128 \times 768$ pixels grayscale image.

\subsection{Imitation learning}\label{sec:imitation-learning}

We use an on-policy imitation learning approach to synthesize a purely vision-based control policy, denoted by $\learner$, that matches the behavior of the position-based flocking policy, denoted by $\expert$, as closely as possible.
More formally, we denote the learned policy $\learner(\obs_t) = \action_t$ as a mapping from observations to actions, where the observations $\obs_t \in \Real^{128 \times 768}$ are grayscale images and the actions $\action_t \in \Real^3$ velocity commands for each time step $t \in T$.
The expert policy $\expert(\state_t) = \action_t$, on the other hand, computes velocity commands from the state $\state_t$ of the system, which is represented by known relative positions $\rel_{ij}$ to other agents as described in Sec.~\ref{sec:flocking-algorithm}.
The image observations can be seen as a lossy representation of the underlying system state (e.g., the relative positions of other agents) because of adverse factors such as limited resolution, occlusions, lens distortions, and inherent noise in the system.
To learn the vision-based policy $\learner$, we use the \textsc{DAgger} imitation learning algorithm \cite{ross_reduction_2011} (see Alg.~\ref{alg:dagger}).

\begin{algorithm}[t]
    \DontPrintSemicolon
    \BlankLine
    Initialize empty dataset $\mathcal{D} \leftarrow \emptyset$.\;
    Initialize parameters of learned policy $\learner_1$.\;
    \For{$i \leftarrow 1 $ \KwTo $N$}{
        Sample trajectories from learned policy $\learner_i$ \tit{simultaneously for all agents}.\;
        Collect dataset $\dataset_i = \{ (\obs_t, \expert(\state_t)) \}_{t = 1}^T$ of observations from the learned policy $\learner_i$ and actions given by expert $\expert$ \tit{for all agents}.\;
        Aggregate datasets $\dataset \leftarrow \dataset \cup \dataset_i$.\;
        Train new policy $\learner_{i + 1}$ on $\dataset$.\;
    }
    \Return{\tno{Best policy $\learner_i$ on hold-out validation set $\dataset_{\mtx{val}}$}}.\;
    \caption{Multi-agent dataset aggregation.}
    \label{alg:dagger}
\end{algorithm}

We collect data and train our policy in an iterative fashion, first in simulation and then in our motion tracking hall.
We use a roughly $80\%/20\%$ split between training $\dataset_\mtx{train}$ and validation data $\dataset_\mtx{val}$.
In simulation, each iteration of the imitation learning algorithm proceeds as follows.
The drones take off and assume random positions within a cube of side length $4 \si{\meter}$, and with a minimum inter-agent distance of $1.5 \si{\meter}$.
The side length and minimum distance were chosen to resemble a plausible real-world deployment scenario in a confined environment such as our motion tracking hall.
All agents then switch to vision-based control and use raw velocity commands generated by the learned policy (which is randomly initialized at first) from the visual inputs sampled at $10 \si{\hertz}$.
Simultaneously, ground truth control commands are computed from the flocking algorithm and stored for post-processing.
The iteration is considered complete as soon as 1) any two drones collide, 2) any two drones are too far away from each other, or 3) $200$ observation-action samples are generated.
We consider two agents too close if any pair of drones falls below a collision threshold of $1 \si{\meter}$; similarly, we consider two agents as too far away when the distance between them exceeds a threshold of $7 \si{\meter}$.
The collision threshold follows the constraints of the drone model, and the dispersion threshold stems from the diminishing size of other agents in the field of view.
For data collection in the real world, we relax the above requirements and stop an iteration as soon as the situation becomes subjectively too dangerous, for instance when the inter-agent distance becomes too small, or the drone starts to move to close to the walls of the motion tracking hall.
A new policy is then trained using the collected image samples, the control commands generated by the learned policy, and the expert control commands computed from the flocking algorithm rules.
Finally, the data collection process is repeated with the new policy.

\subsection{Domain adaptation}\label{sec:domain-adaptation}

One fundamental problem with the on-policy imitation learning approach outlined in Alg.~\ref{alg:dagger} is that the policy needs to be executed in the real world in order to collect samples.
Executing an untrained policy in a multi-agent setting with quadcopters in a confined space such as a motion tracking hall can be dangerous.
A possible solution is to set an initial policy $\policy_i(\obs_t) = \beta_i \expert(\state_t) + (1 - \beta_i) \learner_i(\obs_t)$, i.e. a linear combination of the expert and learner's action and let the factor $\beta_i$ decay from one to zero over time.
While this approach would work, data generation in the real world is error-prone and collecting large datasets would require significant amounts of time.

\begin{figure}[t]
\centering
\begin{subfigure}[b]{\columnwidth}
    \centering
    \frame{\includegraphics[width=\textwidth]{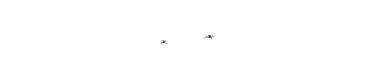}}
    \caption{Foreground}\label{fig:foreground}
\end{subfigure}
\par\smallskip
\begin{subfigure}[b]{\columnwidth}
    \centering
    \frame{\includegraphics[width=\textwidth]{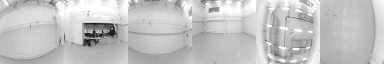}}
    \caption{Background}\label{fig:background}
\end{subfigure}
\par\smallskip
\begin{subfigure}[b]{\columnwidth}
    \centering
    \frame{\includegraphics[width=\textwidth]{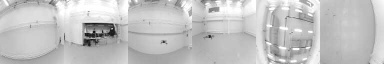}}
    \caption{Fake sample (foreground + background)}\label{fig:fake-sample}
\end{subfigure}
\par\smallskip
\begin{subfigure}[b]{\columnwidth}
    \centering
    \frame{\includegraphics[width=\textwidth]{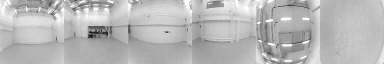}}
    \caption{Real sample}\label{fig:real-sample}
\end{subfigure}
\caption{%
    Example of unsupervised domain adaptation method in which simulated foreground images (\ref{fig:foreground}) and real background images from our motion tracking hall (\ref{fig:background}) are combined into domain-adapted images (\ref{fig:fake-sample}).
    For visual comparison, we also show a real sample from a two-agent flight in the motion tracking hall (\ref{fig:real-sample}).
    }\label{fig:domain-adaptation}
\end{figure}

To avoid the collection of a large real-world dataset, we propose a simple yet effective task-specific domain adaptation method to learn an initial vision-based policy from simulated and unsupervised images.
To this end, we construct a simulated environment in which there is no visual clutter such that the drones appear in front of a uniform white background (see Fig.~\ref{fig:foreground}).
Next, we collect a $20k$-sample image background dataset from our six onboard cameras during a single-agent flight in our motion tracking hall (see Fig.~\ref{fig:background}).
During the flight, we tried to rotate the drone in all possible orientations and to cover as much of the space in the hall as possible to increase the variability in the image data.
As a final step, we add the simulated drones onto the background in order to create a dataset that resembles actual drones flying in the motion tracking hall.
The real and domain-adapted images are almost indistinguishable to the human eye at the resolution used by the control policy (see Fig.~\ref{fig:fake-sample} and \ref{fig:real-sample}).
The control actions corresponding to the images from the simulated dataset remain unchanged during this process.

\subsection{Visual policy}\label{sec:visual-policy}

We formulate the vision-based imitation of the flocking algorithm as a regression problem which takes an image (see Fig.~\ref{fig:field-of-view}) as an input and predicts a velocity command which matches the ground truth velocity command as closely as possible.
To produce the desired velocities, we consider a small and efficient convolutional neural network \cite{loquercio_dronet_2018} that is geared towards drone navigation.
However, unlike \cite{loquercio_dronet_2018}, we opt for a single-head regression architecture to avoid convergence problems caused by different gradient magnitudes from an additional classification objective during training.
This simplifies the optimization problem and the model architecture and thus the resulting controller.

We use mini-batch stochastic gradient descent to minimize the regularized mean squared error loss between predicted and target velocity commands.
We employ variance-preserving parameter initialization by drawing the initial weights from a truncated normal distribution according to \cite{he_delving_2015}.
The biases of the model are initialized to zero.
The objective function is minimized using the Adam optimizer \cite{kingma_adam_2014} and an initial learning rate of $10^{-3}$ which is decayed by a factor of $0.5$ after $10$ consecutive epochs without improvement on the hold-out validation set.
We train the network using a mini-batch size of $128$, a weight decay factor of $5 \cdot 10^{-4}$, and a dropout probability of $0.5$.
We stop the training process as soon as the validation loss plateaus for more than ten consecutive epochs.

The raw images and velocity targets are pre-processed using feature standardization such that each input batch has a mean of zero and a standard deviation of one.
For the velocity targets from the flocking algorithm, we perform a frame transformation from the world frame $\World$ into the drone's body frame $\Body$ as $\target{\vel_i} = {\Rotation^\Body_\World}_i \vel^\txt{rey}_i$ where ${\Rotation_\World^\Body}_i \in \SO{3}$ denotes the rotation matrix from world to body frame for robot $i$ and $\vel^\txt{rey}_i$ corresponds to the target velocity command.
We perform the inverse rotation to transform the predicted velocity commands from the neural network back into the world frame.
In terms of data augmentation, we randomly adjust the image brightness and contrast of each mini-batch by $\pm 25 \%$.
Furthermore, we randomly rotate the image cube map and the control command in $90^\circ$ increments around the body frame z-axis (yaw) such that the vision-based controller becomes invariant to the direction in which agents are predominantly present in the data.
In practice, this is equivalent to shifting and wrapping around the first four images (left, front, right, and back) in $128$-pixel increments, as well as rotating the last two images (top and bottom) by $90^\circ$ increments.


\section{Simulation Results}\label{sec:simulation-results}

\newcommand{\cwidth}{0.04\textwidth}
\newcommand{\fwidth}{0.45\textwidth}

\begin{figure*}[t]
    \begin{subfigure}{\cwidth}
        \caption{}\label{fig:common-position}
    \end{subfigure}
    \begin{subfigure}{\fwidth}
        \includegraphics[width=\textwidth]{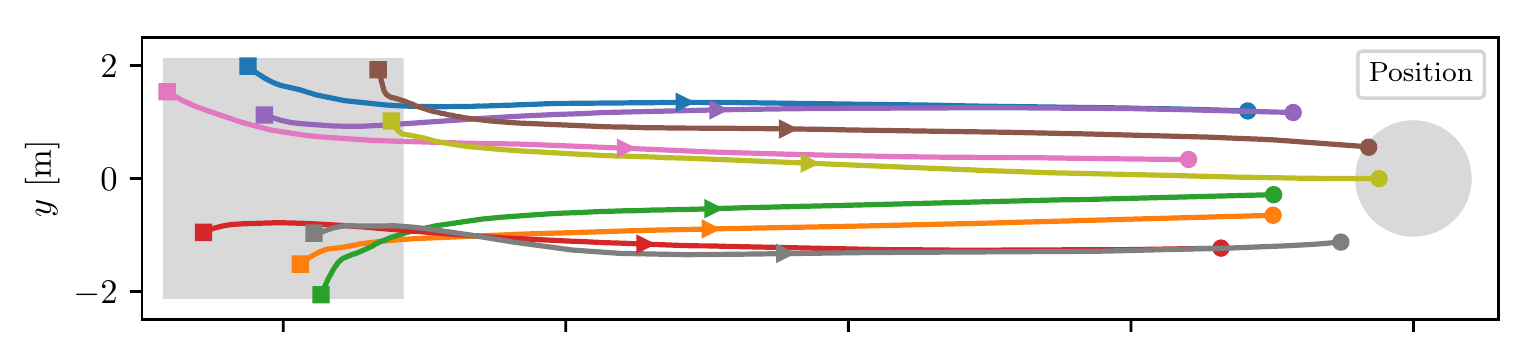}
    \end{subfigure}
    \hfill
    \begin{subfigure}{\fwidth}
        \includegraphics[width=\textwidth]{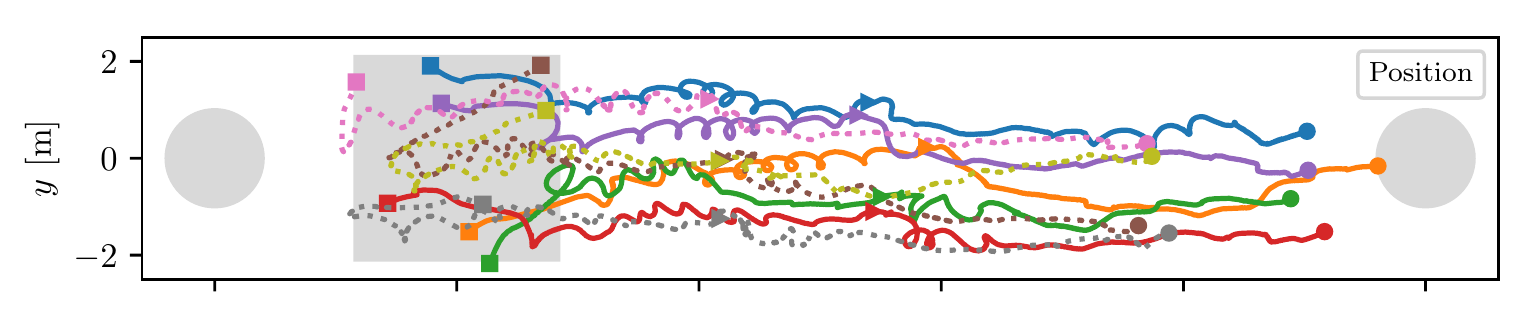}
    \end{subfigure}
    \begin{subfigure}{\cwidth}
        \caption{}\label{fig:opposing-position}
    \end{subfigure}
    \\
    \begin{subfigure}{\cwidth}
        \caption{}\label{fig:common-vision}
    \end{subfigure}
    \begin{subfigure}{\fwidth}
        \includegraphics[width=\textwidth]{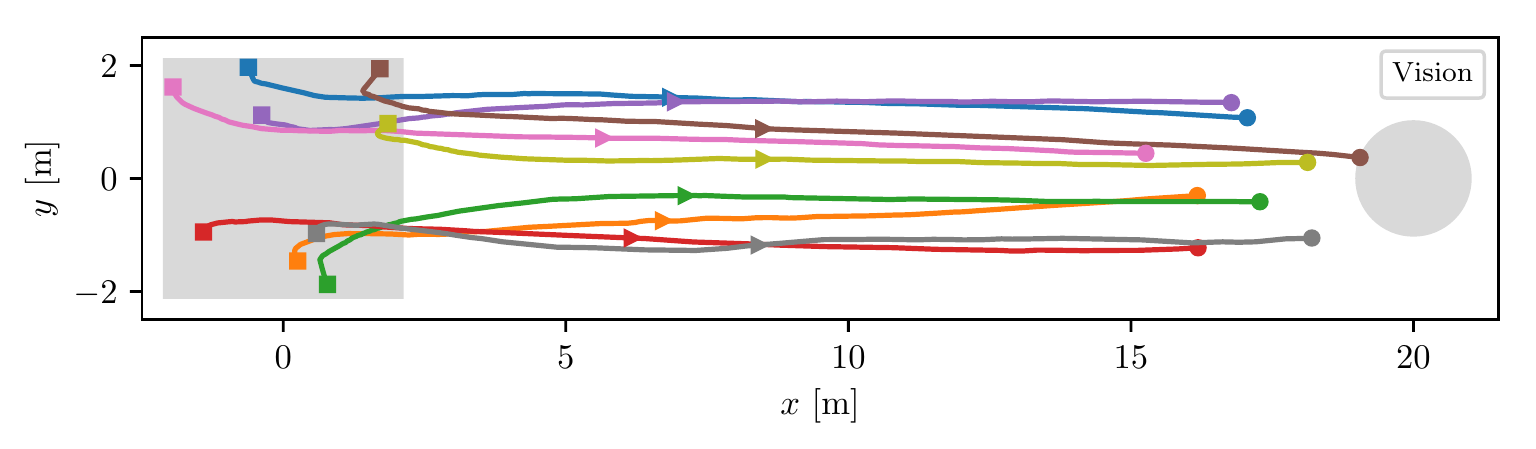}
    \end{subfigure}
    \hfill
    \begin{subfigure}{\fwidth}
        \includegraphics[width=\textwidth]{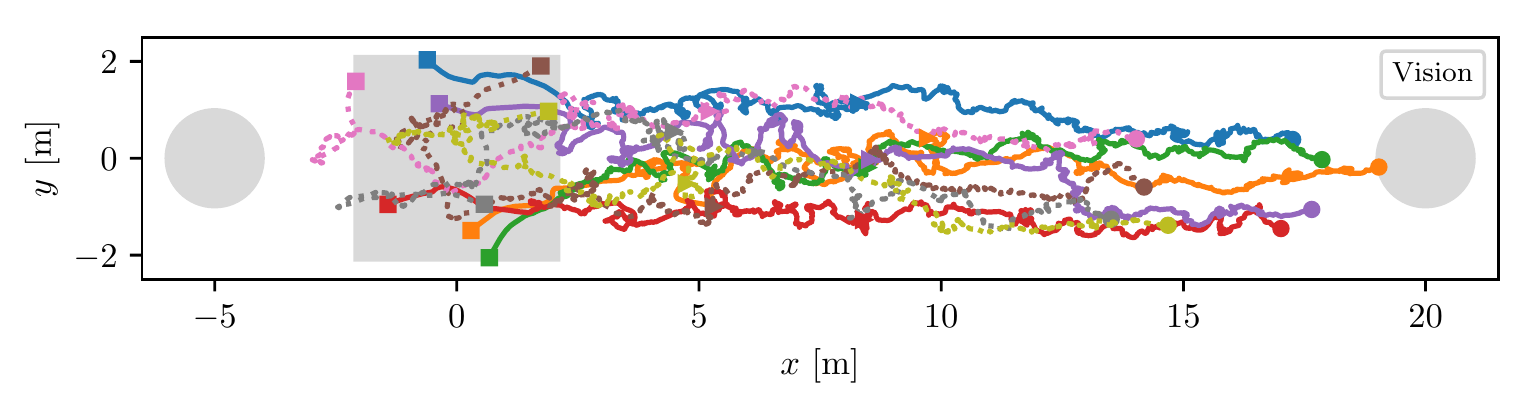}
    \end{subfigure}
    \begin{subfigure}{\cwidth}
        \caption{}\label{fig:opposing-vision}
    \end{subfigure}
    \begin{subfigure}{\cwidth}
        \caption{}\label{fig:common-distances}
    \end{subfigure}
    \begin{subfigure}{\fwidth}
        \includegraphics[width=\textwidth]{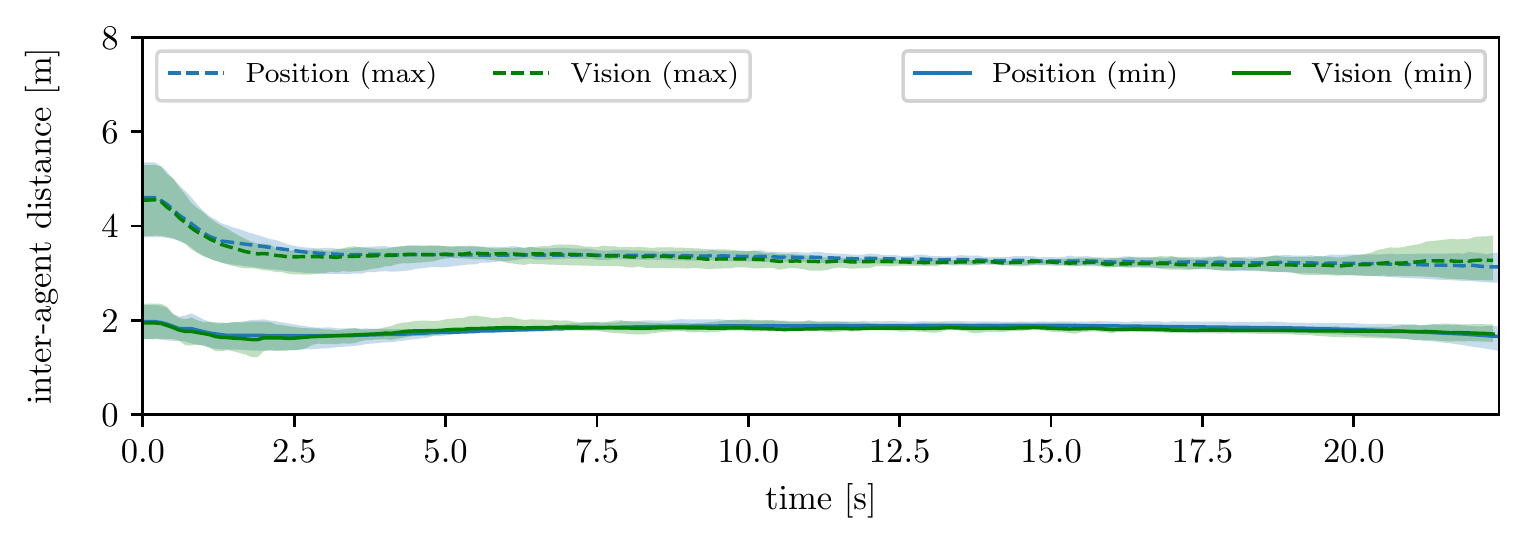}
    \end{subfigure}
    \hfill
    \begin{subfigure}{\fwidth}
        \includegraphics[width=\textwidth]{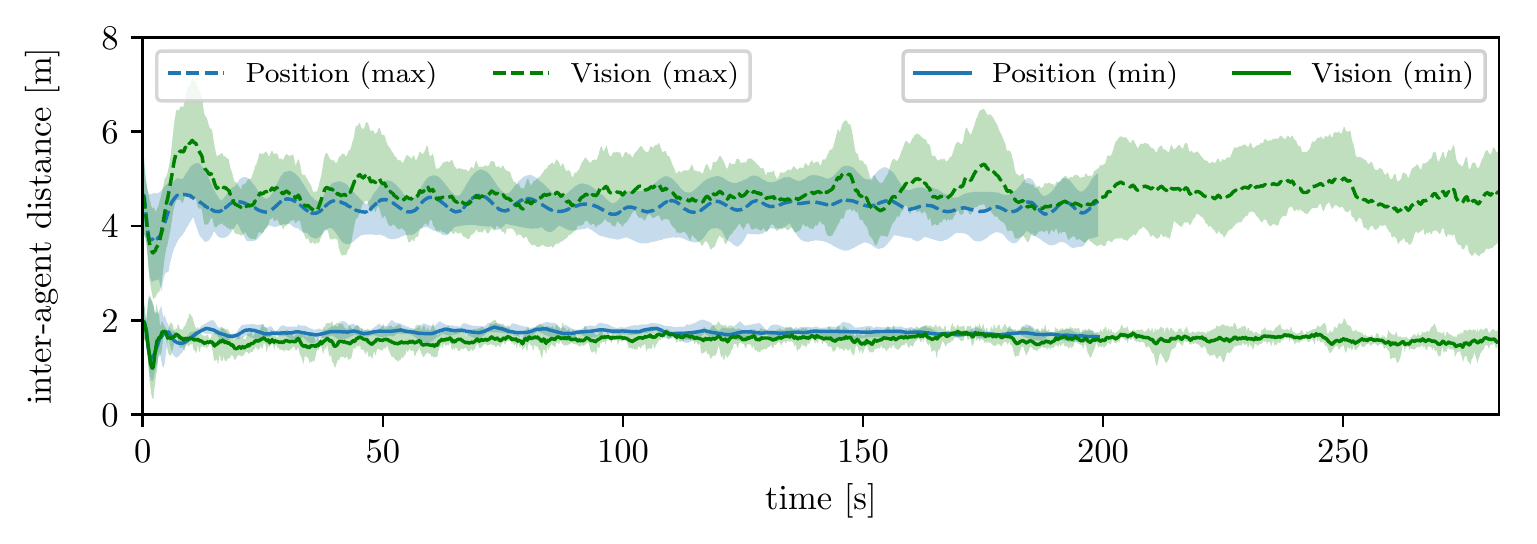}
    \end{subfigure}
    \begin{subfigure}{\cwidth}
        \caption{}\label{fig:opposing-distances}
    \end{subfigure}
    \caption{%
        Flocking with common migration goal (left column) and opposing migration goals (right column).
        \textbf{First two rows}: Top view of a swarm migrating using the \tit{position-based} (\ref{fig:common-position} and \ref{fig:opposing-position}) and \tit{vision-based} (\ref{fig:common-vision} and \ref{fig:opposing-vision}) controller.
        The trajectory of each agent is shown in a different color.
        The colored squares, triangles, and circles show the agent positions during the first, middle, and last time step, respectively.
        The gray square and gray circle denote the spawn area and the migration point, respectively.
        For the swarm with opposing migration goal (\ref{fig:opposing-position} and \ref{fig:opposing-vision}), the waypoint on the right is given to a subset of five agents (solid lines), whereas the waypoint on the left is given to a subset of four agents (dotted lines).
        \textbf{Third row}: Inter-agent minimum and maximum distances over time (\ref{fig:common-distances} and \ref{fig:opposing-distances}) while using the \tit{position-based} and \tit{vision-based} controller.
        The mean minimum distance between any pair of agents is denoted by a solid line, whereas mean maximum distances are shown as a dashed line.
        The colored shaded regions show the minimum and maximum distance between any pair of agents.
        Note that the plot for the opposing goal swarm does not continue until the last time step since the \tit{vision-based} swarm takes longer than the \tit{position-based} swarm to reach the migration point.
    }\label{fig:migration-trajectories}
\end{figure*}

This section presents an evaluation of the learned controller as a comparison to the target flocking algorithm.
We refer to the swarm operating on the learned controller (which relies on visual inputs) as \tit{vision-based}.
We refer to the swarm operating on the flocking algorithm (which relies on shared agent positions) as \tit{position-based}.
The results show that the proposed controller represents a robust alternative to communication-based systems in which the positions of other agents are shared with other members of the group.

The experiments are performed using the Gazebo simulator in combination with the PX4 autopilot \cite{meier_px4_2015} for state estimation and control.
The neural network is implemented in PyTorch.
We employ the same set of flocking parameters used during the training phase throughout the following experiments.
We set the number of agents $N = 9$, the maximum perception radius $r^\txt{max} = 7 \si{\m}$, and the maximum speed $v^\txt{max} = 2 \si{\m\per\s}$.
We set the separation, cohesion, and migration gain to $k^\txt{sep} = 7$, $k^\txt{coh} = 1$, and $k^\txt{mig} = 1$, respectively.

We report our results in terms of minimum and maximum inter-agent distances, two complementary metrics that describe the state of the swarm at a given time step.
The minimum and maximum inter-agent distance are direct indicators for successful collision avoidance, as well as general segregation of the swarm, respectively.
Two conditions are tested: a first one in which all agents share a common migration goal, and a second one in which a subset of the agents have an opposing migration goal.

\subsection{Common migration goal experiment}\label{sec:common-migration-goal-experiment}

In the first experiment, we give all agents the same migration goal and show that the swarm remains collision-free during navigation.
The \tit{vision-based} and the \tit{position-based} swarm exhibit remarkably similar behavior while migrating (see Figs.~\ref{fig:common-position} and \ref{fig:common-vision}).
For the \tit{vision-based} controller, one should notice that the velocity commands predicted by the neural network are sent to the agents in their raw form without any further processing.
The \tit{vision-based} swarm matches the \tit{position-based} one very well since the inter-agent distances do not deviate significantly over the course of the entire trajectory (see Fig.~\ref{fig:common-distances}).
The minimum inter-agent distance remains larger than the collision threshold of $1 \si{\m}$, which indicates that the neural controller has learned to keep a minimum inter-agent distance and thus to avoid collisions.

\subsection{Opposing migration goals experiment}\label{sec:opposing-migration-goals-experiment}

In this experiment, we assign different migration goals to two subsets of agents.
The first group, consisting of five agents, is assigned the same waypoint as in Sec.~\ref{sec:common-migration-goal-experiment}.
The second group, consisting of the remaining four agents, is assigned a migration point on the opposite side with respect to the first group.
The \tit{position-based} and \tit{vision-based} swarm exhibit very similar migration behaviors (see Figs.~\ref{fig:opposing-position} and \ref{fig:opposing-vision}).
In both cases, the swarm cohesion is strong enough to keep the agents together despite the diverging migration goals.
Note that the \tit{vision-based} swarm reaches its migration goal far later than the \tit{position-based} swarm.


\section{Real-world Results}\label{sec:real-world-results}

We propose three experiments involving two quadrotors to show that the learned controller can perform vision-based markerless flight in the real world.
We conclude with an attribution study that visualizes the regions of the visual input that contribute the most to the neural network's predictions.
All flights are performed in our motion tracking hall that is equipped with $26$ OptiTrack cameras.
Each drone receives its ground truth pose via Wi-Fi at a frequency of $\SI{100}{\hertz}$.

\subsection{Circle experiment}\label{sec:circle-experiment}

\begin{figure}
    \centering
    \begin{subfigure}[t]{0.49\columnwidth}
        \includegraphics[width=\textwidth]{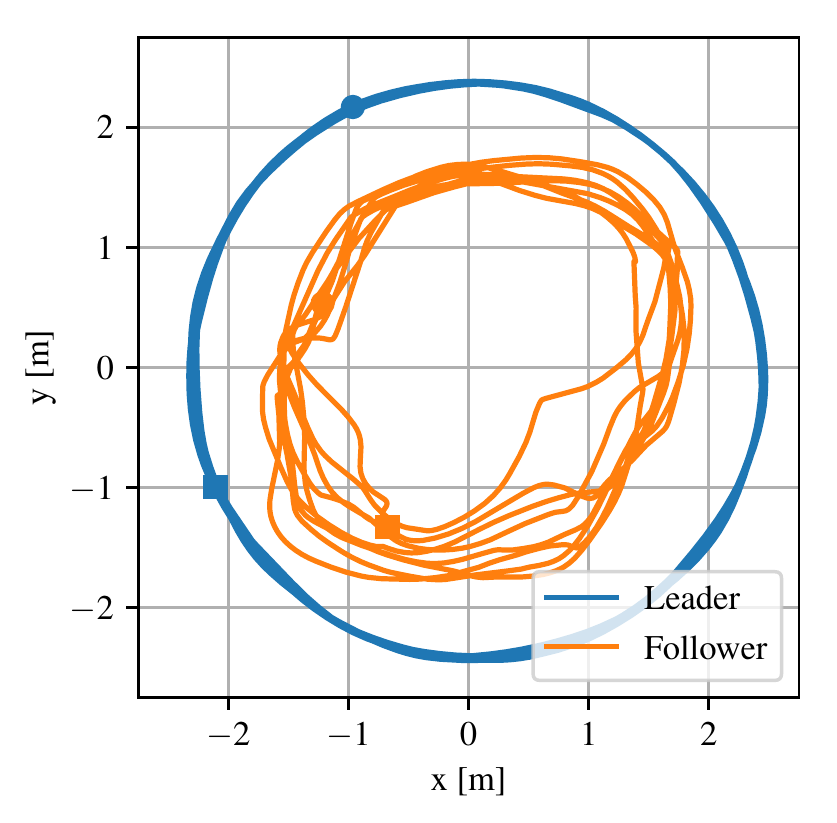}
        \caption{Trajectories (top view)}\label{fig:circle-top}
    \end{subfigure}
    \begin{subfigure}[t]{0.49\columnwidth}
        \includegraphics[width=0.94\textwidth]{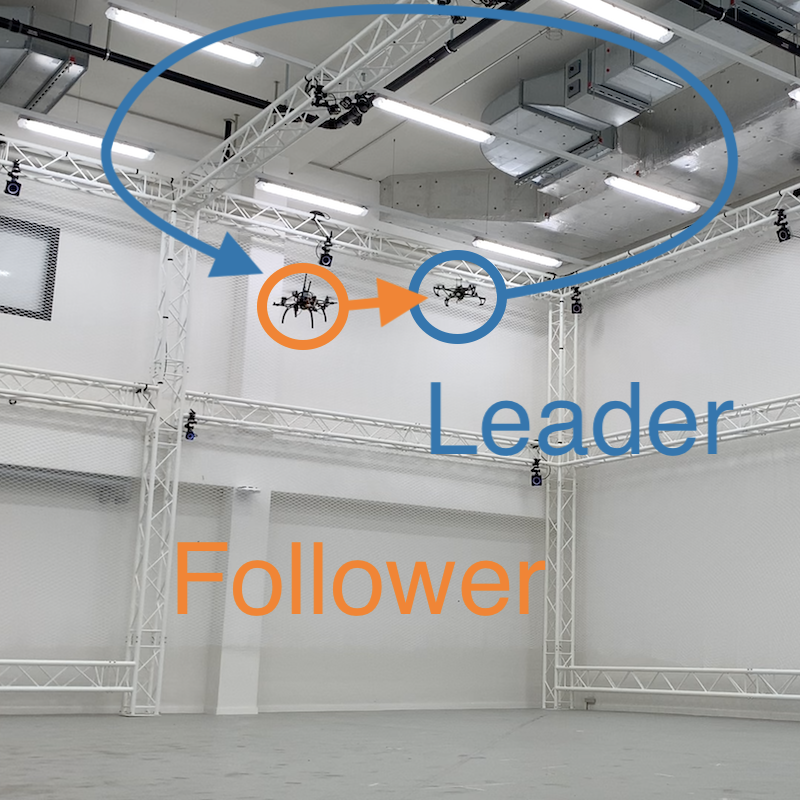}
        \caption{Experimental setup}\label{fig:circle-3d}
    \end{subfigure}
    \begin{subfigure}{\columnwidth}
        \includegraphics[width=\textwidth]{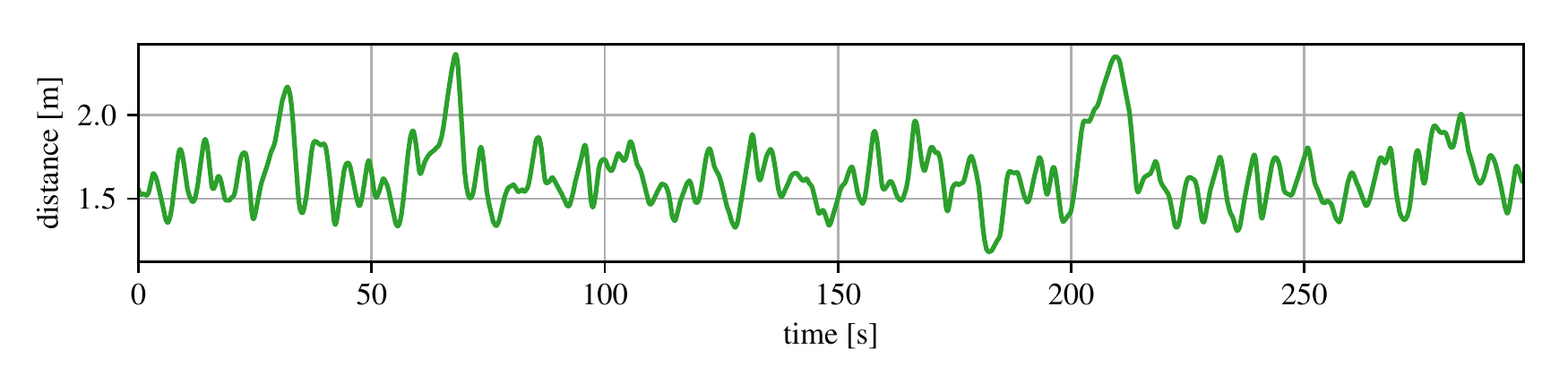}
        \caption{Inter-agent distance over time}\label{fig:circle-distance}
    \end{subfigure}
    \caption{Ground truth trajectories and inter-agent distance during the circle experiment.}\label{fig:circle}
\end{figure}

The circle experiment showcases the ability of the learned policy to maintain cohesion with another agent.
To this end, the leader drone is given a circular trajectory, whereas the vision-based follower uses raw velocity commands generated onboard by the neural network.
We set the radius of the circle as $\SI{2.5}{\meter}$ and the angular velocity along the circular trajectory to $\SI{10}{\degree\per\second}$.
The follower drone keeps a stable distance between itself and the leader drone during a representative $\SI{6}{\minute}$ flight (see Fig.~\ref{fig:circle}).
One can observe that the visual policy can recover from small mistakes reliably, most notably after the $\SI{70}{\second}$ and $\SI{210}{\second}$ marks (see Fig.~\ref{fig:circle-distance}).

\subsection{Carousel experiment}\label{sec:carousel-experiment}

\begin{figure}
    \centering
    \begin{subfigure}{0.49\columnwidth}
        \includegraphics[width=\textwidth]{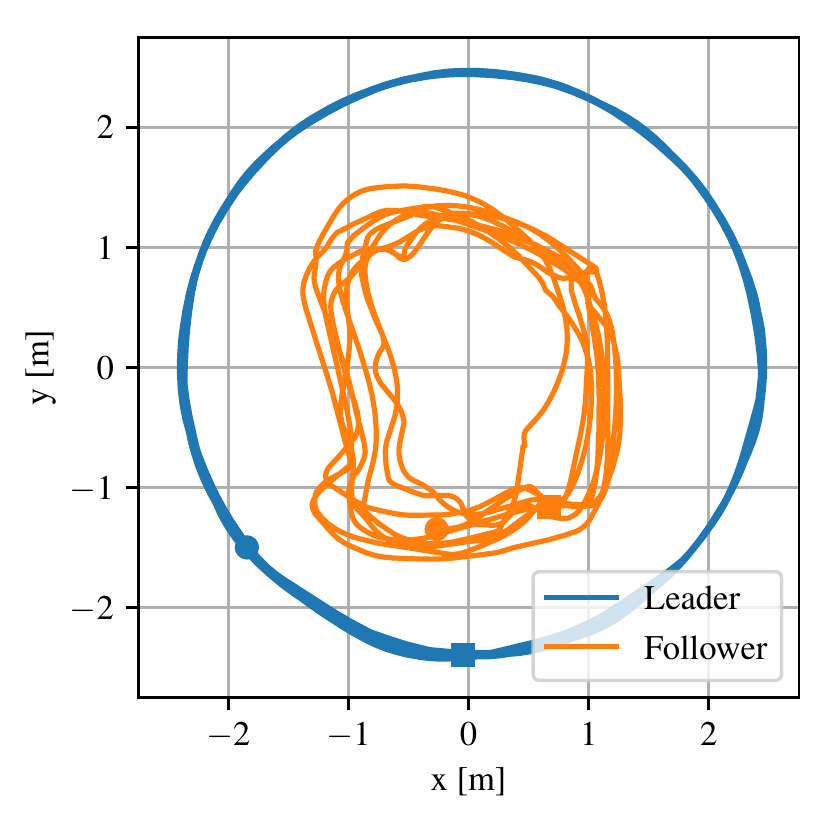}
        \caption{Trajectories (top view)}\label{fig:carousel-top}
    \end{subfigure}
    \begin{subfigure}{0.49\columnwidth}
        \includegraphics[width=\textwidth]{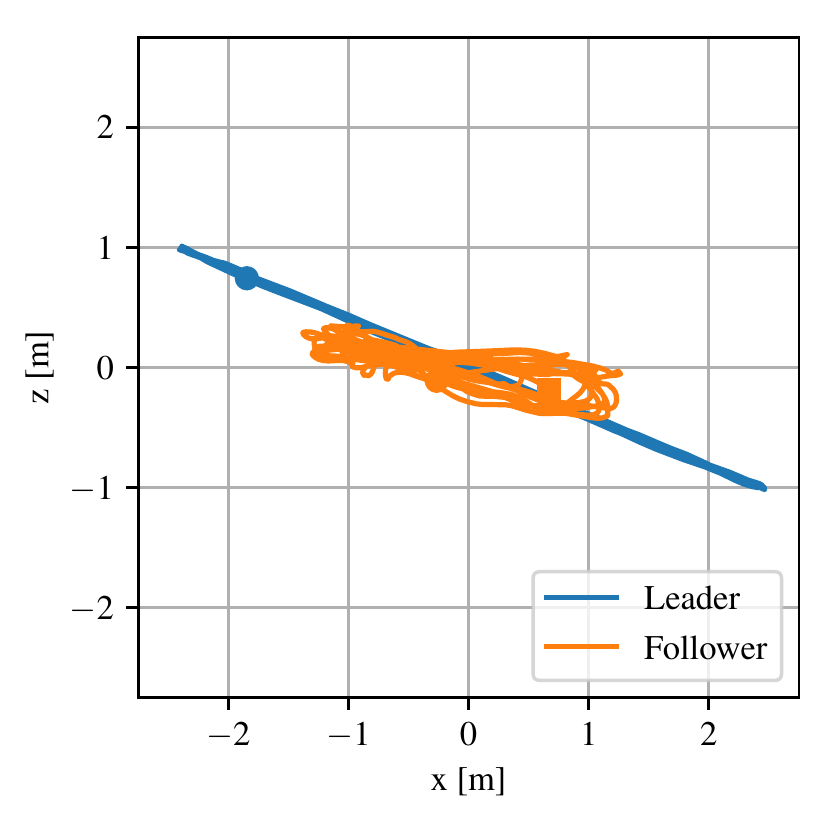}
        \caption{Trajectories (side view)}\label{fig:carousel-side}
    \end{subfigure}
    \begin{subfigure}{\columnwidth}
        \includegraphics[width=\textwidth]{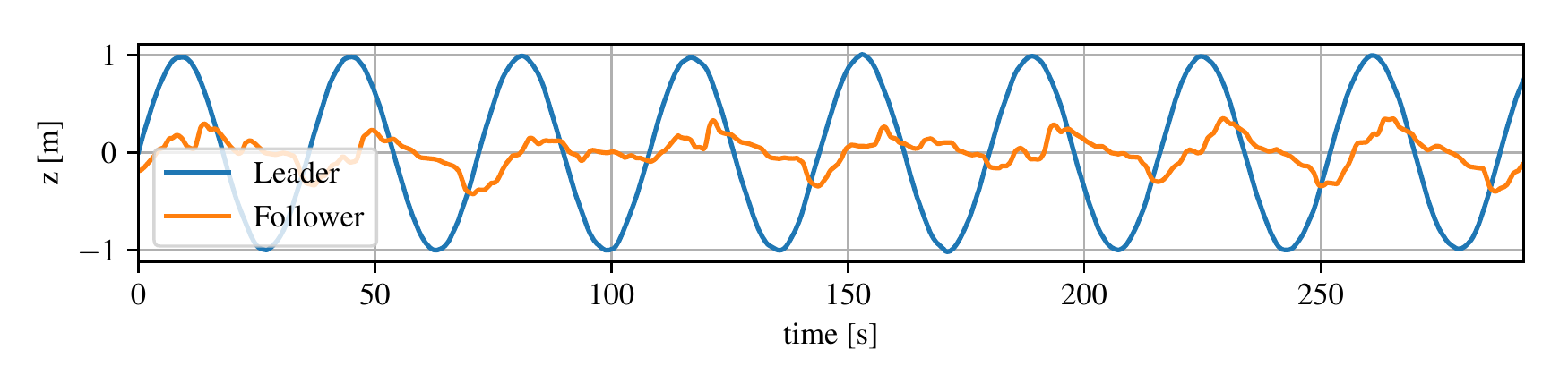}
        \caption{Altitude over time}\label{fig:carousel-altitude}
    \end{subfigure}
    \caption{Ground truth trajectories and altitude during the carousel experiment.}\label{fig:carousel}
\end{figure}

The carousel experiment can be seen as an extension of the circle scenario with the added difficulty that the altitude of the leader is now modulated by a sinusoid as well.
The parameters of the circle trajectory remain the same, but we add a sinusoid component to the altitude tracked by the leader drone.
The altitude component has an amplitude of $\SI{1}{\meter}$ and the same frequency as the horizontal components, which leads to a tilted circular trajectory (see Fig.~\ref{fig:carousel-side}).
The vision-based follower thus needs the ability to operate in full 3D space in order to stay cohesive with the leader.
The inter-agent distance in the carousel experiment increases slightly compared to the circle experiment, especially when the leader agent deviates the most from the average flight altitude.
Nevertheless, the vision-based drone can maintain a steady cohesion with the leader (see Fig.~\ref{fig:carousel}).
Upon closer examination of the altitude of both agents over time, it is clear that the follower can modulate its altitude as a reaction to the leader.
The extreme points in altitude of the follower are indeed somewhat aligned with the intersection points of the two curves (see Fig.~\ref{fig:carousel-altitude}).

\subsection{Push-pull experiment}\label{sec:push-pull-experiment}

\begin{figure}
    \centering
    \begin{subfigure}[t]{0.49\columnwidth}
        \includegraphics[width=\textwidth]{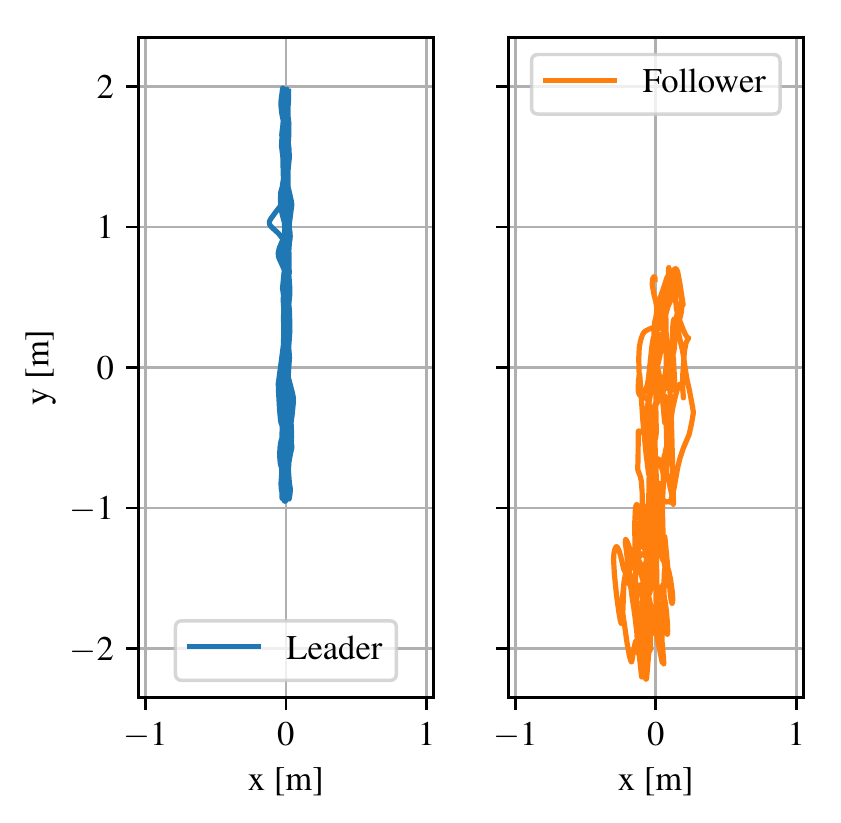}
        \caption{Trajectories (top view)}\label{fig:push-pull-top}
    \end{subfigure}
    \begin{subfigure}[t]{0.49\columnwidth}
        \includegraphics[width=0.94\textwidth]{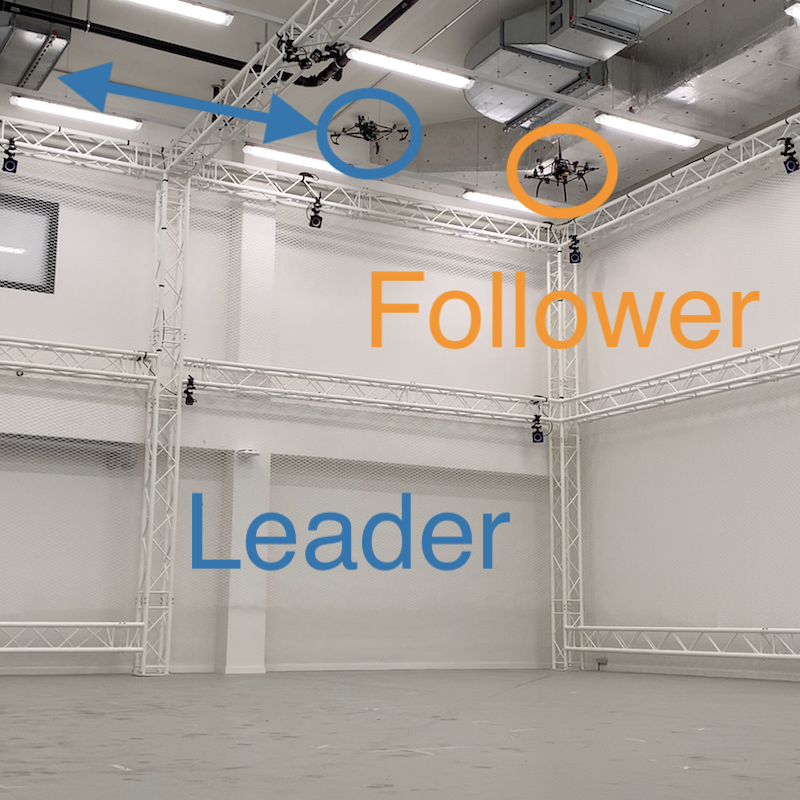}
        \caption{Experimental setup}\label{fig:push-pull-photo}
    \end{subfigure}
    \begin{subfigure}{\columnwidth}
        \includegraphics[width=\textwidth]{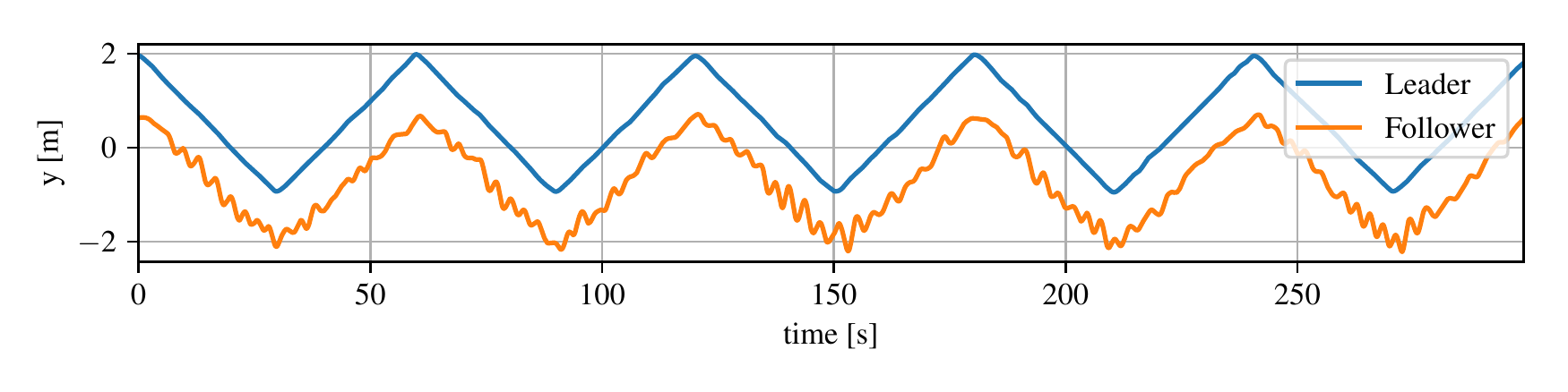}
        \caption{Positions over time}\label{fig:push-pull-position}
    \end{subfigure}
    \caption{Ground truth trajectories and positions during the push-pull experiment.}\label{fig:push-pull}
\end{figure}

The motivation for the push-pull experiment is the validation of the separation ability of the vision-based flocking policy.
In both the circle and carousel experiments, the follower drone is never on a direct collision course with the leader.
In order to encourage collisions, we let the leader navigate between two waypoints and position the follower in the middle.
We further set both the $x$ and the $z$-component of the velocity command computed by the neural network to zero in order to fix the follower's degrees of freedom to a line defined by the two waypoints.
This adjustment is necessary to show collision avoidance since leaving the control input unrestricted would degenerate into a cohesion-like scenario where collisions are not explicitly encouraged.
Moreover, since the drones may get into situations where they are on top of each other, downwash may blur the lines between the separation due to the learned controller and the physical repulsion due to the airflow.
The vision-based follower avoids collisions and maintains a constant equilibrium distance to the leader drone (see Fig.~\ref{fig:push-pull}).
This behavior results directly from the spring-like dynamics of two agents in which one is following the flocking algorithm.
The oscillations in the position may occur because distances smaller than equilibrium are penalized quadratically, while distances larger than equilibrium are penalized linearly by the flocking rules (see Eq.~\ref{eq:separation-cohesion}).
In addition, noise in the raw control command leads to small and sudden repulsive maneuvers that are quickly compensated (see Fig.~\ref{fig:push-pull-position}).

\subsection{Attribution study}\label{sec:attribution-study}

\begin{figure}
    \centering
    \includegraphics[width=\columnwidth]{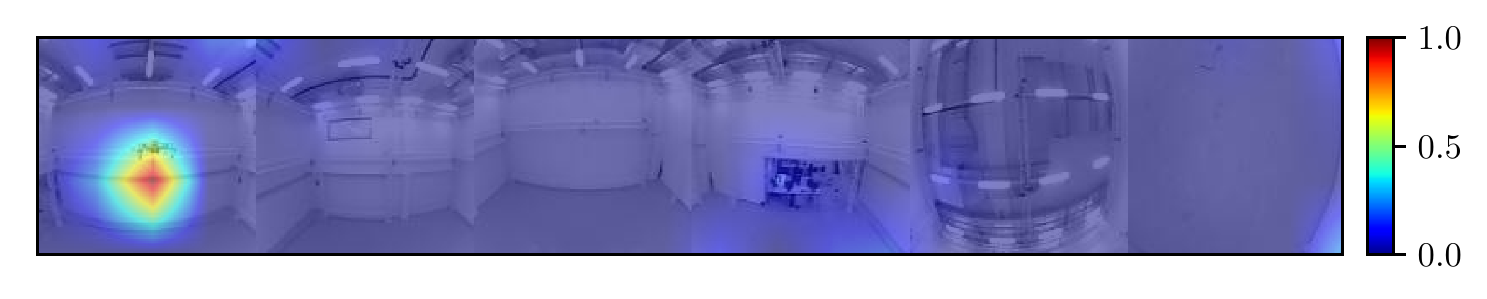}
    \caption{%
        Heat map visualization of the relative importance of each pixel in the visual input of the drone towards its velocity command.
        Red regions have the most influence on the control command, whereas blue regions contribute the least.
        Best viewed in color.
    }\label{fig:heatmap}
\end{figure}

Since the \tit{vision-based} controller provides a very tight coupling between perception and control, the need for interpretation of the learned behavior arises.
To this end, we employ a state-of-the-art attribution method \cite{selvaraju_grad-cam_2017}, which shows how much influence each pixel in the input image has on the predicted velocity command (see Fig.~\ref{fig:heatmap}).
More specifically, we compute the gradients for the heat map with respect to the last convolutional layer of the neural network in which the individual feature maps have a spatial size of only $8 \times 48$ pixels.
We then employ bilinear upsampling to increase the resolution of the resulting saliency map before we blend it with the original input image using a jet colormap for visualization purposes.
The attribution map can be generated very efficiently using one forward and backward pass and could therefore serve as a valuable attention-like input for further real-time processing.

One can observe that the network is effectively localizing the other agent spatially in the visual input.
However, the network is putting non-zero importance on regions with more visual clutter such as the control room in the backward-facing camera (see Fig.~\ref{fig:heatmap}).
One may note that the most salient region is not perfectly matching the location of the visible agent, which can be attributed to the low spatial resolution of the activations generated at the last convolutional layer.

\section{Conclusions and Future Work}\label{sec:conclusions}

This paper presented a machine learning approach to the problem of collision-free and coherent motion of a dense swarm of quadcopters.
The agents learn to coordinate themselves entirely via visual inputs in 3D space by mimicking a flocking algorithm.
The learned controller removes the need for communication of positions among agents and thus presents the first step towards a fully decentralized vision-based swarm of drones.
The trajectories of the swarm are relatively smooth even though the controller is based on raw neural network predictions.
Our algorithm naturally handles navigation tasks by adding a migration term to the predicted velocity of the neural controller.

Regarding future work, a natural subsequent step will be to scale up the real-world experiments with more vision-based drones, as well as the transfer of the learned controller to outdoor scenarios where ground truth positions will be obtained using RTK-capable GNSS receivers.
To reduce the need for large amounts of labeled data, we are exploring recent advances in unsupervised domain adaptation to aid generalization of the neural controller to environments with background clutter.
Another challenge is the addition of obstacles to the environment in which the agents operate.


\section*{Acknowledgments}\label{sec:acknowledgments}

We thank Olexandr Gudozhnik, Vivek Ramachandran, and Enrica Soria for their valuable contributions.

\bibliographystyle{IEEEtran}
\bibliography{library/strings/conferences-abrv,library/strings/journals-abrv,library/library}

\begin{thebibliography}{10}
\providecommand{\url}[1]{#1}
\csname url@samestyle\endcsname
\providecommand{\newblock}{\relax}
\providecommand{\bibinfo}[2]{#2}
\providecommand{\BIBentrySTDinterwordspacing}{\spaceskip=0pt\relax}
\providecommand{\BIBentryALTinterwordstretchfactor}{4}
\providecommand{\BIBentryALTinterwordspacing}{\spaceskip=\fontdimen2\font plus
\BIBentryALTinterwordstretchfactor\fontdimen3\font minus
  \fontdimen4\font\relax}
\providecommand{\BIBforeignlanguage}[2]{{%
\expandafter\ifx\csname l@#1\endcsname\relax
\typeout{** WARNING: IEEEtran.bst: No hyphenation pattern has been}%
\typeout{** loaded for the language `#1'. Using the pattern for}%
\typeout{** the default language instead.}%
\else
\language=\csname l@#1\endcsname
\fi
#2}}
\providecommand{\BIBdecl}{\relax}
\BIBdecl

\bibitem{floreano_science_2015}
D.~Floreano and R.~J. Wood, ``Science, technology and the future of small
  autonomous drones,'' \emph{Nature}, vol. 521, no. 7553, pp. 460--466, 2015.

\bibitem{zufferey_aerial_2011}
J.-C. Zufferey, {Hauert, Sabine}, {Stirling, Timothy}, {Leven, Severin},
  {Roberts, James}, and {Floreano, Dario}, ``Aerial {{Collective Systems}},''
  in \emph{Handbook of {{Collective Robotics}}}, S.~Kernbach, Ed.\hskip 1em
  plus 0.5em minus 0.4em\relax {Pan Stanford}, 2011, pp. 609--660.

\bibitem{kushleyev_towards_2013}
A.~Kushleyev, D.~Mellinger, C.~Powers, and V.~Kumar, ``Towards a swarm of agile
  micro quadrotors,'' \emph{Auton Robots}, vol.~35, no.~4, pp. 287--300, 2013.

\bibitem{preiss_crazyswarm_2017}
J.~A. Preiss, W.~Honig, G.~S. Sukhatme, and N.~Ayanian, ``Crazyswarm: {{A}}
  large nano-quadcopter swarm,'' in \emph{Int Conf Rob Autom (ICRA)}, 2017, pp.
  3299--3304.

\bibitem{weinstein_visual_2018}
A.~Weinstein, A.~Cho, G.~Loianno, and V.~Kumar, ``Visual {{Inertial Odometry
  Swarm}}: {{An Autonomous Swarm}} of {{Vision}}-{{Based Quadrotors}},''
  \emph{IEEE Robot Autom Lett (RA-L)}, vol.~3, no.~3, pp. 1801--1807, 2018.

\bibitem{vasarhelyi_outdoor_2014}
G.~Vásárhelyi, C.~Virágh, G.~Somorjai, N.~Tarcai, T.~Szörényi, T.~Nepusz,
  and T.~Vicsek, ``Outdoor flocking and formation flight with autonomous aerial
  robots,'' in \emph{Int Conf Intel Rob Sys (IROS)}.\hskip 1em plus 0.5em minus
  0.4em\relax {IEEE/RSJ}, 2014, pp. 3866--3873.

\bibitem{viragh_flocking_2014}
C.~Virágh, G.~Vásárhelyi, N.~Tarcai, T.~Szörényi, G.~Somorjai, T.~Nepusz,
  and T.~Vicsek, ``Flocking algorithm for autonomous flying robots,''
  \emph{Bioinspir Biomim}, vol.~9, no.~2, p. 025012, 2014.

\bibitem{vasarhelyi_optimized_2018}
G.~Vásárhelyi, C.~Virágh, G.~Somorjai, T.~Nepusz, A.~E. Eiben, and
  T.~Vicsek, ``Optimized flocking of autonomous drones in confined
  environments,'' \emph{Science Robot}, vol.~3, no.~20, p. eaat3536, 2018.

\bibitem{lecun_deep_2015}
Y.~LeCun, Y.~Bengio, and G.~Hinton, ``Deep learning,'' \emph{Nature}, vol. 521,
  no. 7553, pp. 436--444, 2015.

\bibitem{hauert_reynolds_2011}
S.~Hauert, S.~Leven, M.~Varga, F.~Ruini, A.~Cangelosi, J.-C. Zufferey, and
  D.~Floreano, ``Reynolds flocking in reality with fixed-wing robots:
  {{Communication}} range vs. maximum turning rate,'' in \emph{Int Conf Intel
  Rob Sys (IROS)}.\hskip 1em plus 0.5em minus 0.4em\relax {IEEE/RSJ}, 2011, pp.
  5015--5020.

\bibitem{reynolds_flocks_1987}
C.~W. Reynolds, ``Flocks, {{Herds}} and {{Schools}}: {{A Distributed Behavioral
  Model}},'' in \emph{Annual Conf Comp Graph Interactive Technol (SIGGRAPH)},
  vol.~14, 1987, pp. 25--34.

\bibitem{dousse_human-comfortable_2017}
N.~Dousse, G.~Heitz, F.~Schill, and D.~Floreano, ``Human-{{Comfortable
  Collision}}-{{Free Navigation}} for {{Personal Aerial Vehicles}},''
  \emph{IEEE Robot Autom Lett (RA-L)}, vol.~2, no.~1, pp. 358--365, 2017.

\bibitem{ross_learning_2013}
S.~Ross, N.~Melik-Barkhudarov, K.~S. Shankar, A.~Wendel, D.~Dey, J.~A. Bagnell,
  and M.~Hebert, ``Learning {{Monocular Reactive UAV Control}} in {{Cluttered
  Natural Environments}},'' in \emph{Int Conf Rob Autom (ICRA)}, 2013, pp.
  1765--1772.

\bibitem{loquercio_dronet_2018}
A.~Loquercio, A.~I. Maqueda, C.~R. del Blanco, and D.~Scaramuzza, ``{{DroNet}}:
  {{Learning}} to {{Fly}} by {{Driving}},'' \emph{IEEE Robot Autom Lett
  (RA-L)}, vol.~3, no.~2, pp. 1088--1095, 2018.

\bibitem{sadeghi_cad2rl_2017}
F.~Sadeghi and S.~Levine, ``{{CAD2RL}}: {{Real Single}}-{{Image Flight}}
  without a {{Single Real Image}},'' in \emph{Rob: Sci Sys (RSS)}, vol.~13,
  2017.

\bibitem{giusti_machine_2016}
A.~Giusti, J.~Guzzi, D.~C. Cireşan, F.~L. He, J.~P. Rodríguez, F.~Fontana,
  M.~Faessler, C.~Forster, J.~Schmidhuber, G.~D. Caro, D.~Scaramuzza, and L.~M.
  Gambardella, ``A {{Machine Learning Approach}} to {{Visual Perception}} of
  {{Forest Trails}} for {{Mobile Robots}},'' \emph{IEEE Robot Autom Lett
  (RA-L)}, vol.~1, no.~2, pp. 661--667, 2016.

\bibitem{smolyanskiy_toward_2017}
N.~Smolyanskiy, A.~Kamenev, J.~Smith, and S.~Birchfield, ``Toward
  {{Low}}-{{Flying Autonomous MAV Trail Navigation}} using {{Deep Neural
  Networks}} for {{Environmental Awareness}},'' in \emph{Int Conf Intel Rob Sys
  (IROS)}, 2017, pp. 4241--4247.

\bibitem{saska_system_2017}
M.~Saska, T.~Baca, J.~Thomas, J.~Chudoba, L.~Preucil, T.~Krajnik, J.~Faigl,
  G.~Loianno, and V.~Kumar, ``System for deployment of groups of unmanned micro
  aerial vehicles in {{GPS}}-denied environments using onboard visual relative
  localization,'' \emph{Auton Robots}, vol.~41, no.~4, pp. 919--944, 2017.

\bibitem{krajnik_practical_2014}
T.~Krajník, M.~Nitsche, J.~Faigl, P.~Vaněk, M.~Saska, L.~Přeučil,
  T.~Duckett, and M.~Mejail, ``A {{Practical Multirobot Localization
  System}},'' \emph{J Intell Robotic Syst}, vol.~76, no. 3-4, pp. 539--562,
  2014.

\bibitem{walter_uvdar_2019}
V.~Walter, N.~Staub, A.~Franchi, and M.~Saska, ``{{UVDAR System}} for {{Visual
  Relative Localization}} with application to {{Leader}}-{{Follower
  Formations}} of {{Multirotor UAVs}},'' \emph{IEEE Robot Autom Lett (RA-L)},
  pp. 1--1, 2019.

\bibitem{saska_swarms_2014}
M.~Saska, J.~Vakula, and L.~Přeućil, ``Swarms of micro aerial vehicles
  stabilized under a visual relative localization,'' in \emph{Int Conf Rob
  Autom (ICRA)}, 2014, pp. 3570--3575.

\bibitem{soria_influence_2019}
E.~Soria, F.~Schiano, and D.~Floreano, ``The {{Influence}} of {{Limited Visual
  Sensing}} on the {{Reynolds Flocking Algorithm}},'' in \emph{Int Conf Rob
  Comp (IRC)}, 2019, pp. 138--145.

\bibitem{ross_reduction_2011}
S.~Ross, G.~Gordon, and D.~Bagnell, ``A {{Reduction}} of {{Imitation Learning}}
  and {{Structured Prediction}} to {{No}}-{{Regret Online Learning}},'' in
  \emph{Int Conf Artif Intel Stat (AISTATS)}, vol.~14.\hskip 1em plus 0.5em
  minus 0.4em\relax {JMLR.org}, 2011, pp. 627--635.

\bibitem{he_delving_2015}
K.~He, X.~Zhang, S.~Ren, and J.~Sun, ``Delving {{Deep}} into {{Rectifiers}}:
  {{Surpassing Human}}-{{Level Performance}} on {{ImageNet Classification}},''
  in \emph{Int Conf Comp Vis (ICCV)}, 2015, pp. 1026--1034.

\bibitem{kingma_adam_2014}
D.~P. Kingma and J.~Ba, ``Adam: {{A Method}} for {{Stochastic Optimization}},''
  in \emph{Int Conf Learn Repr (ICLR)}, 2014.

\bibitem{meier_px4_2015}
L.~Meier, D.~Honegger, and M.~Pollefeys, ``{{PX4}}: {{A}} node-based
  multithreaded open source robotics framework for deeply embedded platforms,''
  in \emph{Int Conf Rob Autom (ICRA)}, 2015, pp. 6235--6240.

\bibitem{selvaraju_grad-cam_2017}
R.~R. Selvaraju, M.~Cogswell, A.~Das, R.~Vedantam, D.~Parikh, and D.~Batra,
  ``Grad-{{CAM}}: {{Visual Explanations}} from {{Deep Networks}} via
  {{Gradient}}-{{Based Localization}},'' in \emph{Int Conf Comp Vis (ICCV)},
  2017, pp. 618--626.

\end{thebibliography}

\end{document}